Tatjana Scheffler* (Bochum), Veronika Solopova* (TU Berlin), Mihaela Popa-Wyatt (Manchester)
* Erstautorinnen

**Verbreitungsmechanismen schädigender Sprache im Netz: Anatomie zweier Shitstorms**

## 1. Einführung

Der Shitstorm, »eine unvorhergesehene, kurz anhaltende Welle der Empörung in sozialen Medien«[1], findet erst in kürzerer Zeit Betrachtung in der linguistischen Forschung[2]. In diesem Papier widmen wir uns zwei exemplarischen, medienübergreifenden Shitstorms, die sich gegen bekannte Personen aus dem Wirtschaftsleben richten. Beiden gemeinsam ist erstens der Auslöser, eine kontroverse Äußerung der Person, die dadurch zum Ziel des Shitstorms wird, und zweitens die Identität dieser Zielperson als relativ privilegiert: cis-männlich, weiß, erfolgreich. Die Kritik an der auslösenden Äußerung wird zum Shitstorm, als sie in ein zweites virtuelles Medium hinüberschwappt. Wir untersuchen die Ausbreitung der Empörungswelle über jeweils zwei Medien hinweg und testen die Anwendbarkeit von computerlinguistischen Methoden zur Analyse des zeitlichen Verlaufs eines solchen Prozesses. Unter der Annahme, dass sich schädigende Sprache im digitalen Raum wie ein Virus, also durch Ansteckung, verbreitet[3], interessiert uns vor allem, welche Ereignisse und Konstellationen zur Verwendung schädigender Sprache führen, und ob und wie eine linguistische Formierung von »Stämmen«[4] stattfindet. Unsere Forschung konzentriert sich daher erstens auf die Verteilung linguistischer Merkmale innerhalb des gesamten Shitstorms: Werden einzelne Wörter oder Phrasen nach ihrer Einführung verstärkt verwendet und auf welchen Wegen verbreiten sie sich? Zweitens fragen wir, ob sich »Stämme«, zum Beispiel eine Gruppe von Unterstützer*innen und eine von Gegner*innen der Zielperson, linguistisch formieren. Unsere Hypothese ist, dass Unterstützer*innen über den gesamten Zeitverlauf gleichmäßig aktiv bleiben, während der dynamische »Wellen«effekt des Shitstorms auf der unterschiedlichen Beteiligung der Gegner*innen basiert. Final skizzieren wir den zeitlichen Verlauf von plattformübergreifenden Shitstorms schematisch.

---

[1] Gaderer, Rupert (2018): »Shitstorm: Das eigentliche Übel der vernetzten Gesellschaft«, in: Zeitschrift für Medien- und Kulturforschung 9/2, S. 27–42.

[2] Vgl. Bauer, Nathalie et al. (2016): »Streiten 2.0 im Shitstorm–Eine exemplarische Analyse sprachlicher Profilierungsmuster im sozialen Netzwerk Facebook«, in: Reihe XII, S. 157; Bendel, Kay/Menger, Nick/Skottke, Eva-Maria (2016): »Analyse der Wahrnehmung von Shit- und Candystorms mittels Eye-tracking«, in: Hektor Haarkötter (Hg.) Shitstorms und andere Nettigkeiten, Baden-Baden: Nomos, S. 85–108; Gaderer, Rupert (2018): »Shitstorm: Das eigentliche Übel der vernetzten Gesellschaft«, in: Zeitschrift für Medien- und Kulturforschung 9/2, S. 27–42; Haarkötter, Hektor (2016): »Empörungskaskaden und rhetorische Strategien in Shitstorms«, in: Hektor Haarkötter (Hg.) Shitstorms und andere Nettigkeiten, Baden-Baden: Nomos, S. 17–50; Himmelreich, Sascha/Einwiller, Sabine (2015): »Wenn der „Shitstorm“ überschwappt – Eine Analyse digitaler Spillover in der deutschen Print- und Onlineberichterstattung«, in: Olaf Hoffjann/Thomas Pleil (Hg.) Strategische Onlinekommunikation, Wiesbaden: Springer VS, S. 183–205; Kuhlhüser, Sandra (2016): »Shitstorm gleich Shitstorm? – Eine empirische Untersuchung des Netzphänomens exemplarisch dargestellt am Amazon-Shitstorm 2013« in: Hektor Haarkötter (Hg.) Shitstorms und andere Nettigkeiten, Baden-Baden: Nomos, S. 51–84; Marx, Konstanze (2019): »Von Schafen im Wolfspelz. Shitstorms als Symptome einer medialen Emotionskultur«, in: Stefan Hauser/Martin Luginbühl/Susanne Tienken (Hg.) Mediale Emotionskulturen. Bern [u.a.]: Lang, S. 135–153; Marx, Konstanze (2020): »Das Dialogpotenzial von Shitstorms«, in: Ernest W.B. Hess-Lüttich (Hg.) Handbuch Gesprächsrhetorik, Boston: De Gruyter, S. 409–428; Stefanowitsch, Anatol (2020): »Der Shitstorm im Medium Twitter: Eine Fallstudie«, in: Marx, Konstanze/Lobin, Henning/Schmidt, Axel (Hg.) Deutsch in Sozialen Medien: Interaktiv – multimodal – vielfältig, Berlin, Boston: De Gruyter, S. 185–214.

[3] Popa-Wyatt, Mihaela (2023): »Online Hate. Is Hate an Infectious Disease? Is Social Media a Promoter?«, in Journal of Applied Philosopy 40, S. 788–812; Popa-Wyatt, Mihaela (2022): »Social media: a viral promoter of social ills?«, Cardiff University, Internet: https://blogs.cardiff.ac.uk/openfordebate/social-media-a-viral-promoter-of-social-ills/. Zuletzt geprüft am: 29.11.2022.

[4] Deremetz, Anne/Scheffler, Tatjana (2020): »Die Retribalisierung der Gesellschaft?: Transformationen von Twitter-Diskursen zu #DSGVO im Zeitverlauf«, in: Zeitschrift für Kultur- und Kollektivwissenschaft 6/2, S. 171–216.

## 2. Empirische Basis: Zwei Beispiel-Stürme

Als Untersuchungsbasis wählen wir zwei aktuelle Shitstorms, die im September und Oktober 2022 stattfanden. Der erste begann mit einer Twitterumfrage, in der Elon Musk die Unterstützung der Ukraine im Angriffskrieg Russlands in Frage stellte. Der zweite wurde ausgelöst, als der CEO einer eSport-Plattform ein Video twitterte, das ihn beim Feiern mit dem bekannten Misogynisten Andrew Tate zeigte. Alle Daten in den beteiligten Plattformen wurden automatisch mit Python ausgelesen: Telegram mit der Python-Bibliothek Telephon, Reddit mit der Praw-Bibliothek und Twitter mit Tweepy unter Nutzung von akademischen Entwickler-Zugangsdaten.[5]

### 2.1 Elon Musk

Der Shitstorm begann am 3.10., als Elon Musk eine Umfrage zusammen mit seinen Vorschlägen zur Beendigung des Krieges in der Ukraine twitterte (Abb. 1a), sowie später am Tag (Abb. 1b).

Später am selben Tag veröffentlichte der Präsident der Ukraine Wolodymyr Selenskyj einen Tweet mit seiner eigenen Umfrage: »Welchen @elonmusk mögen Sie mehr? Einen, der die Ukraine unterstützt/einen, der Russland unterstützt«[6], wobei 78,8 % von 2,4 Millionen Abstimmende für die erste Option stimmten. Ein weiterer wichtiger Tweet, der schließlich gelöscht wurde, erschien am 6. Oktober. Innerhalb der ukrainischen Gemeinschaft weitete sich der Skandal schnell auf Telegram-Kanäle aus. Wir haben einen davon ausgewertet, nämlich den User-Thread im Blog des ukrainischen Parlamentsabgeordneten Alexij Goncharenko, der sich vom 3. bis zum 6. Oktober ebenfalls mehrfach zu der Situation geäußert hat.

a.
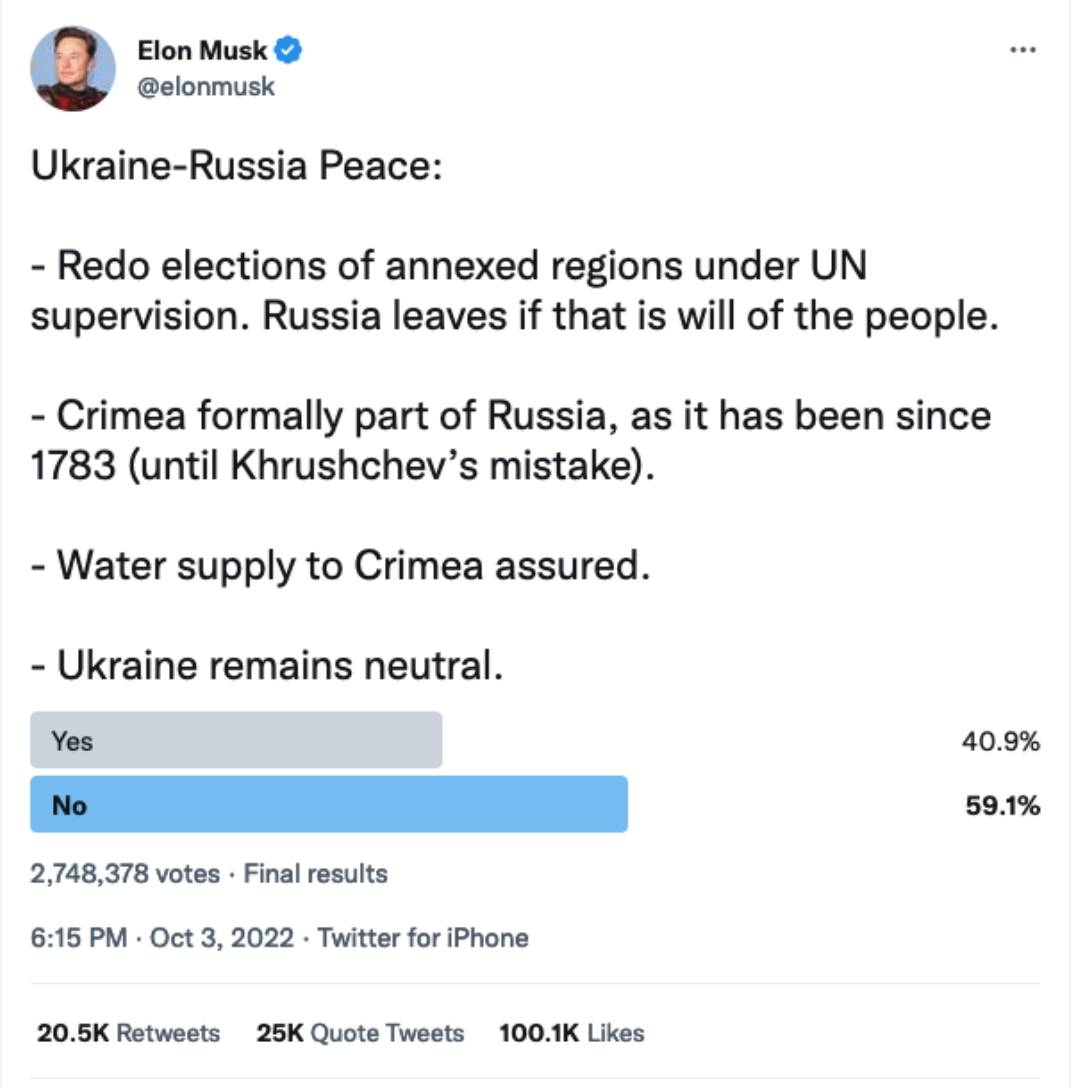

b.
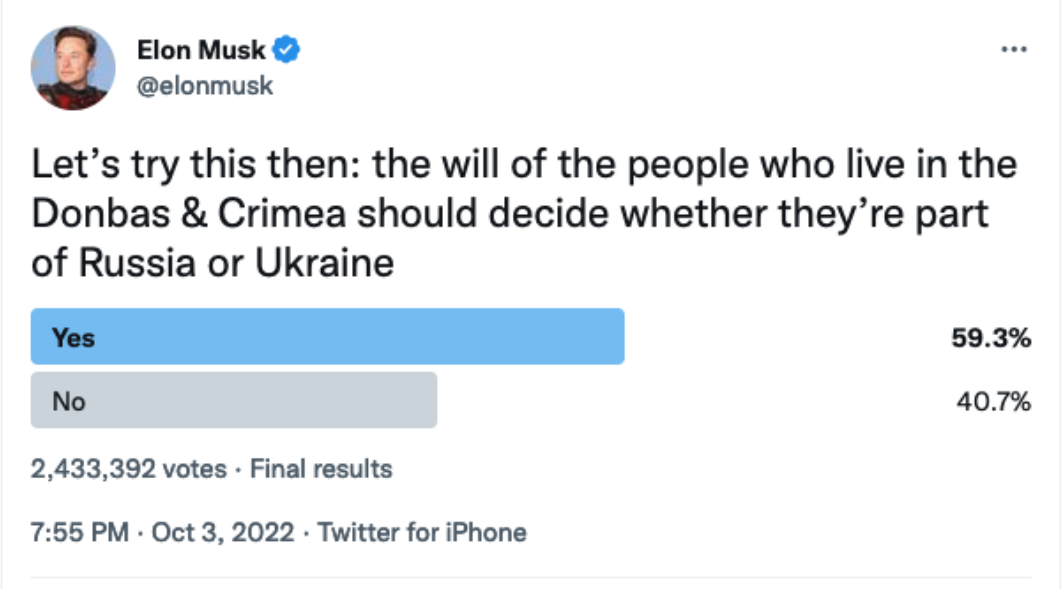

Abbildung 1: Tweets von Elon Musk zum Ukraine-Krieg (3. Oktober 2022), Twitter.

[5] https://docs.telethon.dev, https://praw.readthedocs.io, https://docs.tweepy.org

[6] Alle Übersetzungen der ukrainischen und russischen Posts wurden von uns vorgenommen.

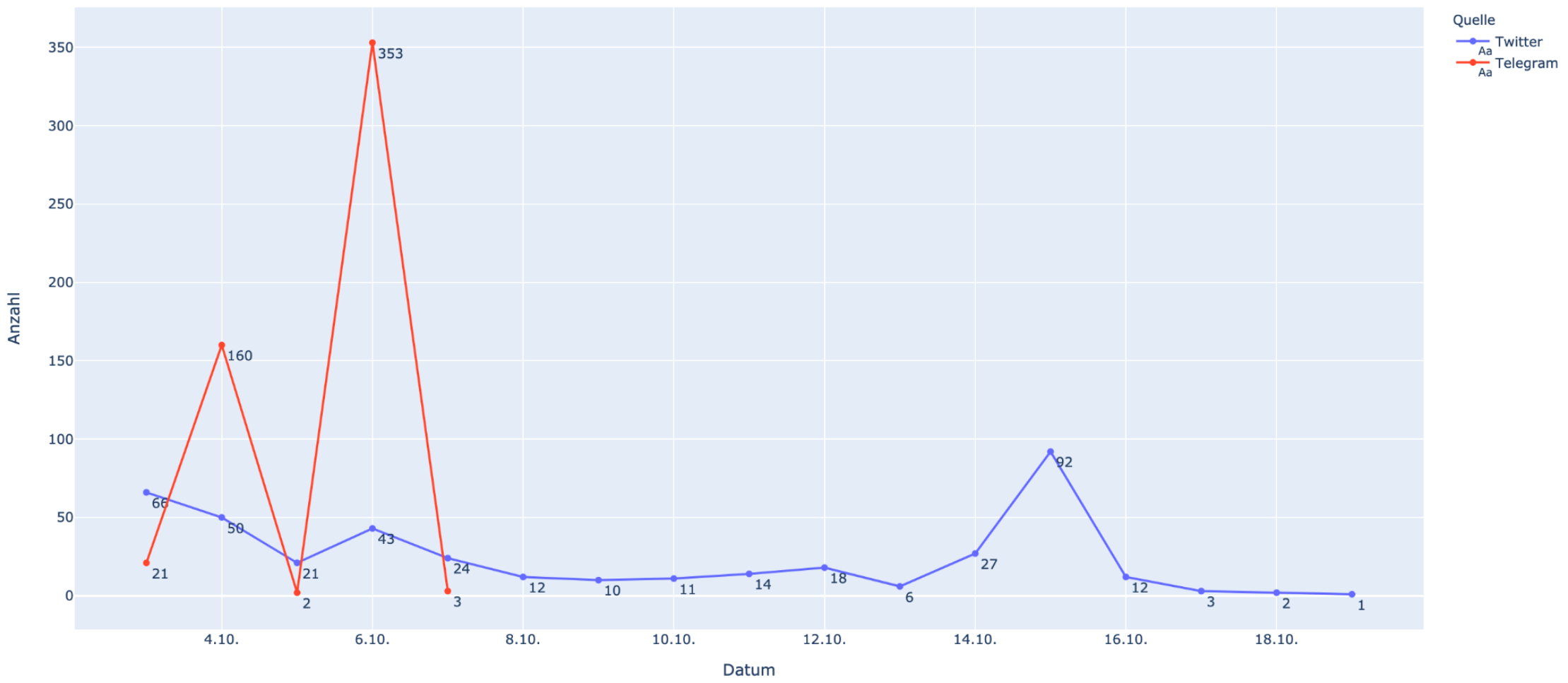

Abbildung 2: Zeitverlauf des Elon-Musk-Shitstorms auf Twitter und Telegram.

Interessant sind vor allem die Hinweise im Telegram-Kanal, in denen ukrainische Leser*innen aufgefordert werden, sich an den Umfragen zu beteiligen und russische Bots abzuwehren: »Da präsentiert sich Elon Musk auf Twitter mal wieder als Experte für Geopolitik. Ich antworte ihm bereits, machen auch Sie mit«, heißt es am 6. Oktober. Diese Einladung ist für die Spitze der Beteiligung an diesem Tag verantwortlich (siehe Abb. 2). Insgesamt wurden 413 Tweets und 539 Nachrichten von Telegram gesammelt.

Im Vergleich der Plattformen haben wir festgestellt, dass die Telegram-Aktivität sehr viel kurzlebiger, aber intensiver war (Abb. 2). Der letzte Telegram-Post zu diesem Thema wurde am 7. Oktober gefunden: »Über Musk. Wissenschaftler haben das Geheimnis von Mona Lisas Lächeln gelöst. Sie ist nur ein Narr.« Auf Twitter ging die Debatte den ganzen Oktober über weiter und endete größtenteils am 19. Oktober, wobei der letzte Tweet von Elon Musk am 15. Oktober erfolgte: »The hell with it … even though Starlink is still losing money & other companies are getting billions of taxpayer $, we'll just keep funding Ukraine govt for free«, was eine Menge dankbarer ukrainischer Tweets nach sich zog.

Nur 8,8 % der Twitter-Diskutant*innen beteiligen sich mehr als einmal an diesen Threads, die Antworten sind hauptsächlich Reaktionen auf den Ursprungstweet und nicht auf eine andere Antwort. Auf Telegram beteiligen sich 28 % der Nutzer*innen mehr als einmal und 13 % mehr als zweimal, was ebenfalls eine intensivere Beteiligung suggeriert.

Die verwendeten Hashtags positionieren sich meist nicht pro Elon Musk, sondern gegen die ukrainische Gemeinschaft, wie z. B. direkt »#standwithrussia« oder »#americanpropaganda« und kontextbezogen wie »#MelnykSeiStill«, »#MelnykShutUp« als Adressierung von Andrij Melnyk, dem ukrainischen Ex-Botschafter in Deutschland, der für sein aktives politisches Engagement in der deutschen Debatte um die Lieferung schwerer Waffen bekannt ist[7]. In den Telegram-Daten fanden wir keine Hinweise auf die Verwendung von Hashtags, da dies weder für die ukrainische Community noch für Telegram als Plattform im Allgemeinen typisch ist[8].

Zur inhaltlichen Charakterisierung der Diskurse bestimmten wir die häufigsten Wörter im Zeitverlauf. Die Schlüsselwörter »Ukraine«, »Russland«, »Frieden«, »Krieg« und »Elon« sind im Kern der Konversation, sie sind vom ersten bis zum letzten Tag, den wir dokumentiert haben, häufig, wobei »Russland« häufiger als jedes andere Schlüsselwort und am 13. und 15. Oktober viel häufiger als das

---

[7] Siehe: https://www.nytimes.com/2022/07/09/world/europe/ukraine-german-ambassador-andriy-melnyk.html

[8] Scheffler, Tatjana/Solopova, Veronika/Popa-Wyatt, Mihaela (2021): »The Telegram Chronicles of Online Harm«, in: Journal of Open Humanities Data 7, S. 8.

Schlüsselwort »Ukraine« vorkommt. Das Wort »unterstützen« bleibt nur bis zum 10. Oktober, während »stoppen« im Sinne von »aufhören, dieses korrupte Land zu finanzieren« nur bis zu dem Tag vorkommt, an dem Musk ankündigt, dass Starlink weiterhin in der Ukraine tätig sein wird, und »wollen« im Sinne von »erklären, was Elon Musk will« beginnt am nächsten Tag der Debatte und setzt sich auch während der gesamten Diskussion fort. Die übrigen Wörter stehen für untergeordnete Themen. So wurde »Selenskyj« vom 4.–9.10. nach seinem Tweet diskutiert, »thank« zeigt die Dankbarkeit der ukrainischen Gemeinschaft; dann wird darüber gesprochen, wie teuer der Starlink-Betrieb in der Ukraine ist, »Tesla« und »Krim« werden nur später diskutiert (s. Anhang A).

Vergleicht man die Twitter- und Telegram-Aktivitäten, so gibt es eine Reihe von Unterschieden, angefangen bei den Emojis. Im Allgemeinen sind Emojis in beiden Datensätzen nicht sehr präsent, jedoch können wir eine eher politische oder unterstützende Tendenz in den Twitter-Daten und spöttische, lachende Emojis häufiger in Telegram sehen (Abb. 3).

Was die beleidigende Sprache betrifft, so ist die Konversation in Telegram viel anfälliger für verletzende Sprache (31 % der Beiträge) als Twitter (5 %), siehe Abb. 4, 5.

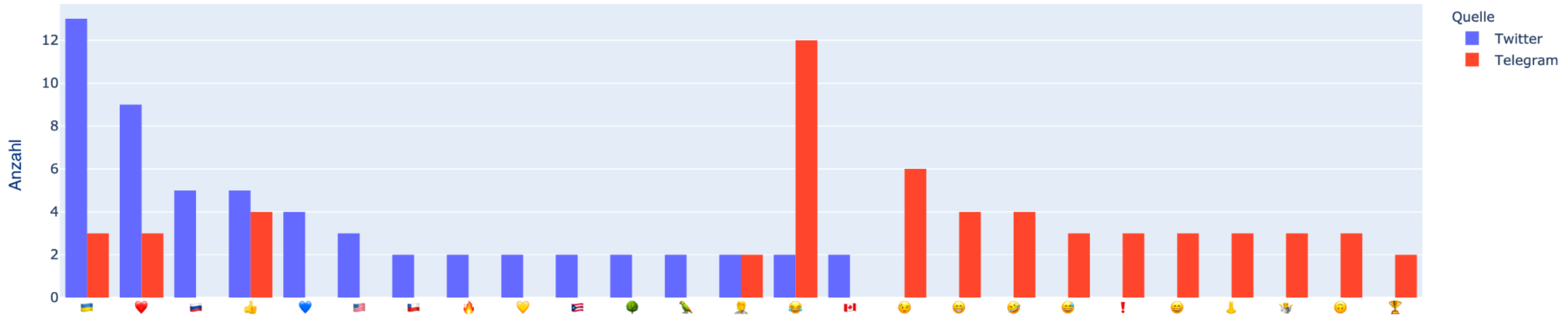

Abbildung 3: Verwendung von Emojis im Shitstorm von Elon Musk, Twitter versus Telegram.

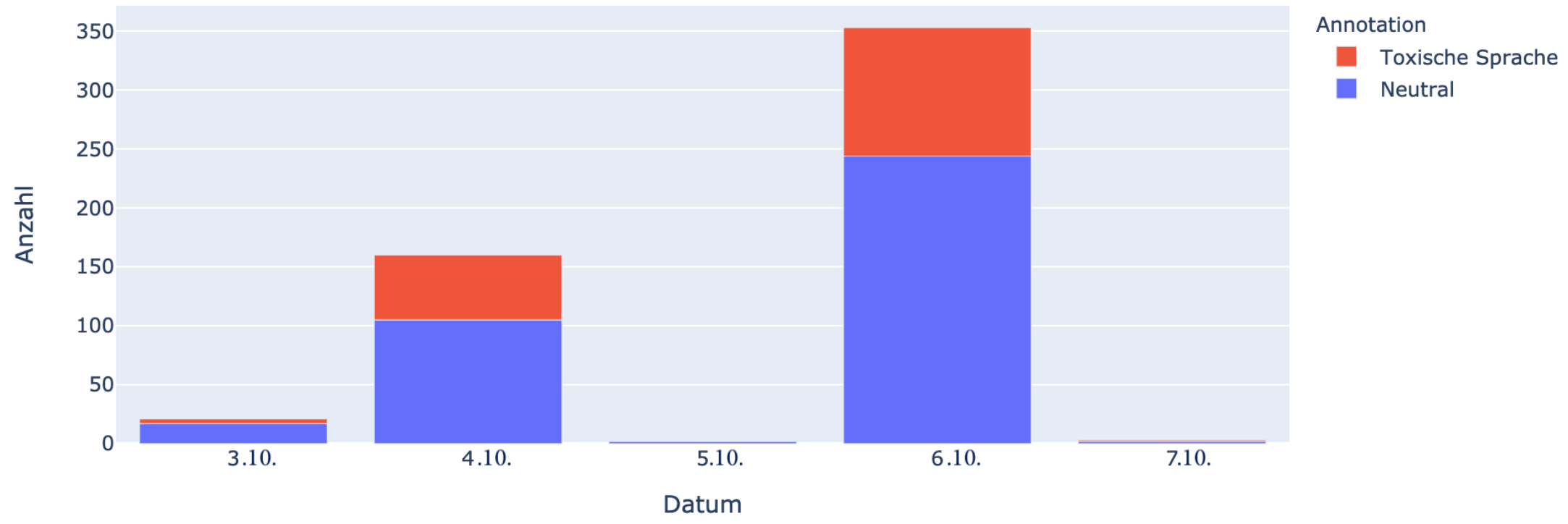

Abbildung 4: Verteilung toxischer Sprache auf Telegram im Shitstorm von Elon Musk (toxische Sprache wurde automatisch erkannt).

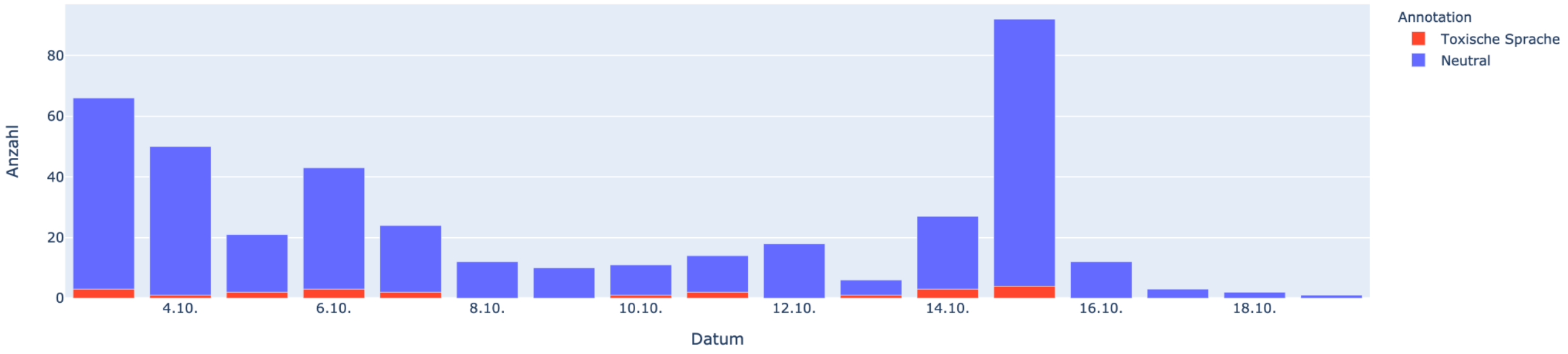

Abbildung 5: Verteilung toxischer Sprache auf Twitter im Shitstorm von Elon Musk.

## 2.2 eSport

Der zweite Shitstorm begann am 17. September, als der CEO von G2 eSports Carlos »Ocelote« ein Video auf Twitter postete, in dem er mit Andrew Tate[9] feiert. Aus der möglicherweise kurzlebigen Empörung wurde ein Shitstorm, als Ocelote seine Unterstützung für Tate bekräftigte und twitterte: »Niemand wird jemals in der Lage sein, meine Freundschaften zu kontrollieren, ich ziehe hier meine Grenze, ich feiere, mit wem ich will«. Später am nächsten Tag wurde zwar eine PR-Erklärung in seinem Namen veröffentlicht, gleichzeitig unterstützte Ocelote aber Tweets, die seine Unschuld unterstützten, durch ein »Like«. Dieser Widerspruch wurde von der Community sehr negativ aufgenommen. Am Abend dieses Tages wurde bekannt gegeben, dass Ocelote als CEO acht Wochen unbezahlten Urlaub nehmen wird, was die Öffentlichkeit jedoch nicht zufrieden stellte. Die Situation eskalierte, bis am 23. September bekannt wurde, dass die Organisation G2 eSports aufgrund der Situation eine sehr wichtige Franchise-Möglichkeit in einem neuen Spiel verloren hatte. Zuletzt war Ocelote gezwungen, die CEO-Position endgültig zu verlassen und sogar seine Aktien an der Firma zu verkaufen (s. Abb. 6). Ähnlich wie im Musk-Shitstorm kamen hier am 18. und 20. September Nutzer*innen von Twitter zu einer anderen Plattform (hier ein eSports-Subreddit), um andere über neue Tweets zu informieren. Sie forderten aber nicht direkt zur Teilnahme am Shitstorm auf.

Während wir bei Reddit keine Nutzer*inneninformationen extrahieren konnten, haben sich bei diesem Twitter-Skandal wieder 7 % der Personen mehr als einmal beteiligt. Emojis sind in den Reddit-Daten nicht sehr präsent, da sie insgesamt nur 13-mal verwendet wurden. Das am häufigsten verwendete Twitter-Emoji ist der Clown, was wiederum die allgemeine Haltung der Community gegenüber der Zielperson zeigt (s. Abb. 7). Auch Hashtags sind nicht typisch für diesen Reddit-Thread: Nur zwei wurden verwendet, eins um sich über die Verwendung von »#MeToo« lustig zu machen, und das kreative Beispiel »#TOPG2«, das den Slogan von Andrew Tate und den Namen des G2-Teams vereint. Auf Twitter wurden Hashtags etwas häufiger verwendet: 2-mal »#Respect«, 2-mal »#G2ARMY«, sowie »#AndrewTate« und einige unerwartete Beispiele wie »#StandWithUkraine«, »#RespawnRecruits«, »#EINS« und »#ABetterABK«, die sich auf eine Vereinigung der Mitarbeiter von Activision Blizzard Kind beziehen.

Die Twitter-Sprache ist in diesem Diskurs insgesamt doppelt so toxisch wie in dem Elon-Musk-Thread mit 10,4 % schädigender Sprache. Auch bei Reddit, der zweiten, mehr auf die eSports-Community ausgerichteten Plattform, gibt es mit 22 % aller Nachrichten mehr schädigende Sprache. Dies zeigt sich auch daran, dass »fuck« eines der häufigsten Schlüsselwörter ist (Abb. 8).

Was die Schlüsselwörter betrifft (Anhang B), so spiegeln die wichtigsten Schlüsselwörter auf beiden Social-Media-Plattformen das Hauptthema wieder: »carlos«, »tate«, »g2«, »people«. Auf Twitter diskutieren die Nutzer*innen später auch darüber, was die wichtigste Seite des Skandals ist: die Party selbst oder der Tweet, in dem der CEO seine Unterstützung für Andrew Tate bekräftigt. Die Diskussion auf Reddit umfasst viele weitere Unterthemen: wie das Image der Organisation beschädigt wurde, wie sich der weibliche Teil der Fangemeinde in diesem Zusammenhang fühlt, etc. Gleichzeitig äußern sich viele Teilnehmer*innen dankbar und entschuldigend Ocelote gegenüber und kritisieren eine angebliche »Cancel culture«, die einem Menschen wegen eines Fehlers das ganze Leben nimmt.

---

[9] Influencer, der auf mehreren Plattformen, darunter Twitter, Instagram und Facebook, gesperrt ist, und der für homophobe, rassistische und frauenfeindliche Kommentare bekannt ist. Gegen ihn wird zum Zeitpunkt der Abfassung dieses Artikels wegen Menschenhandels und Vergewaltigung ermittelt.

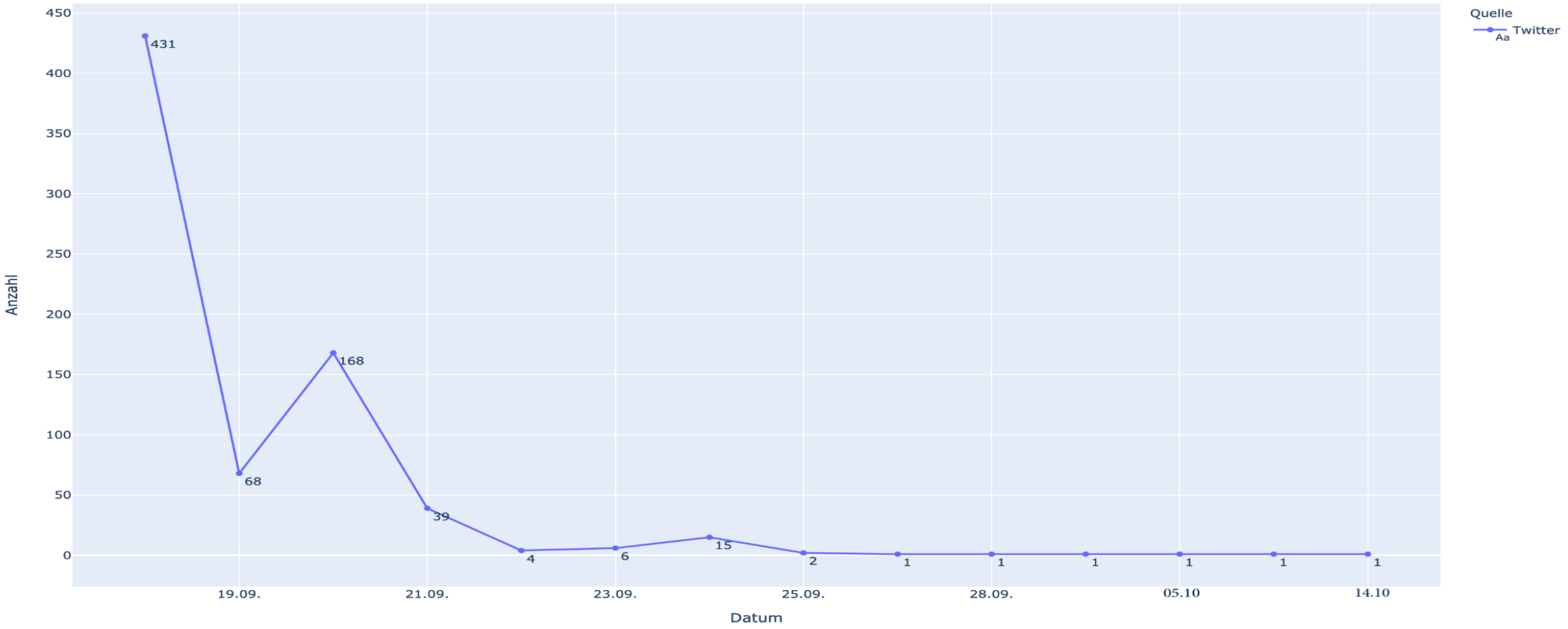

Abbildung 6: Zeitverlauf des eSports-Shitstorms auf Twitter. Genaue Zeitstempel der Redditbeiträge konnten leider nicht aufgerufen werden.

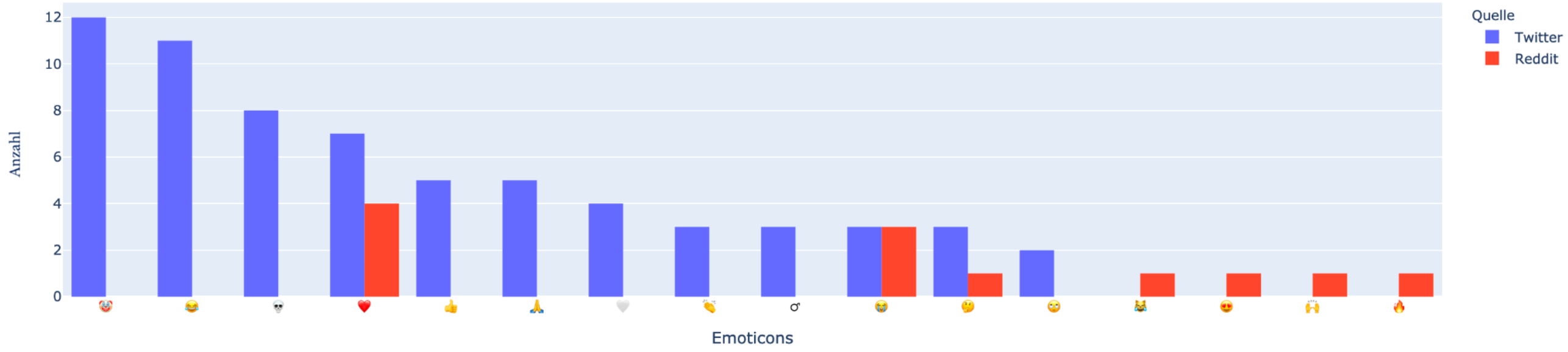

Abbildung 7: Verwendung von Emojis im eSports-Shitstorm, Twitter vs. Reddit.

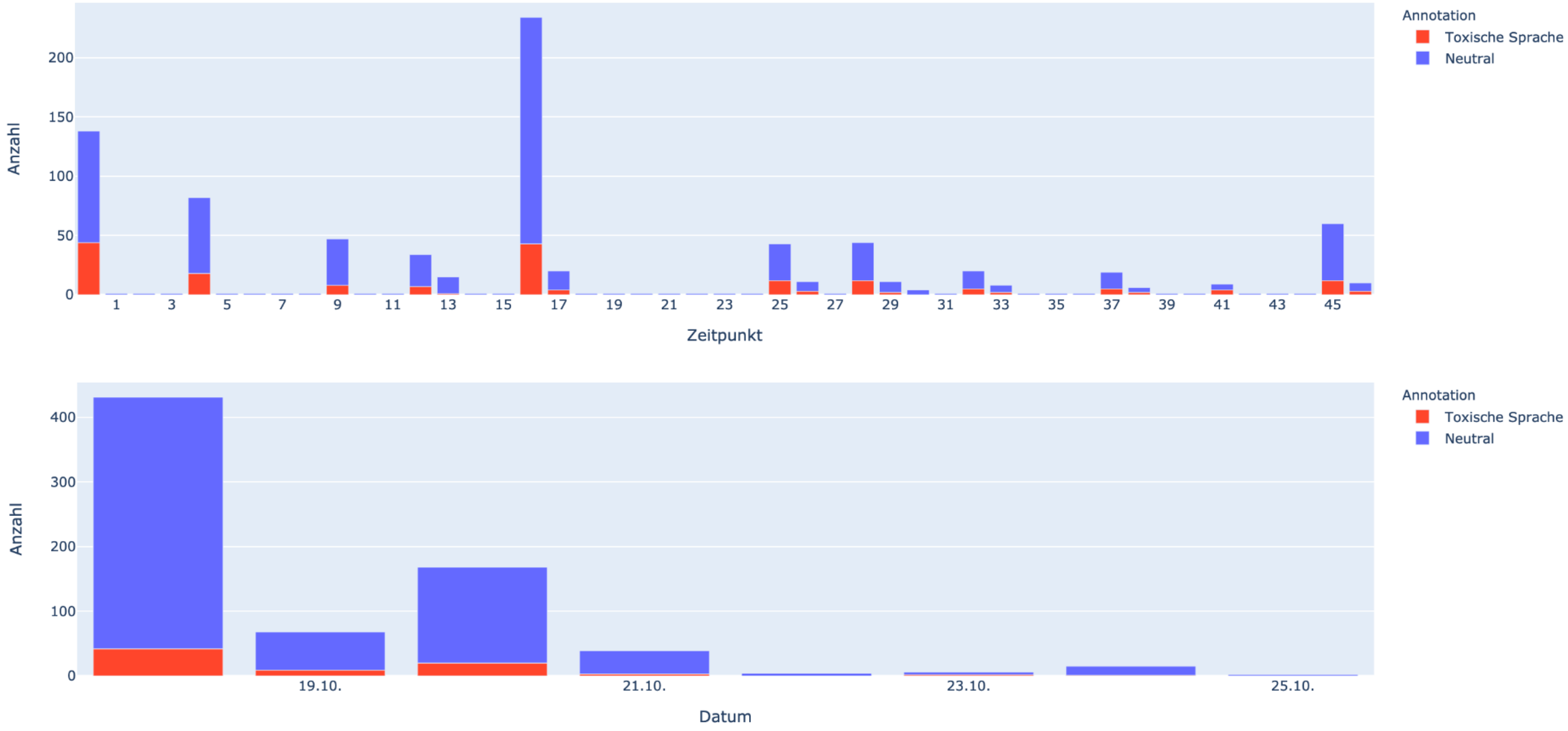

Abbildung 8: Verteilung toxischer Sprache auf Reddit (oben) und Twitter (unten) im eSports-Shitstorm.

## 3. Gruppenbildung: Unterstützer*innen und Gegner*innen im Shitstorm

Der Shitstorm wird von der Konfrontation zwischen Kritiker*innen und Unterstützer*innen der auslösenden Person getragen. Zentral für die Konstitution der »Welle« ist die kontinuierliche Reibung zwischen Für und Wider. Im Elon-Musk-Shitstorm konnten wir die plattformübergreifende

Rekrutierung des Wider-Lagers explizit beobachten. Wir analysieren daher die einzelnen Beiträge der beiden Diskurse, um Unterstützer*innen und Gegner*innen zu finden. Dies verspricht Aufschluss über den Verlauf der Shitstorms zu geben.

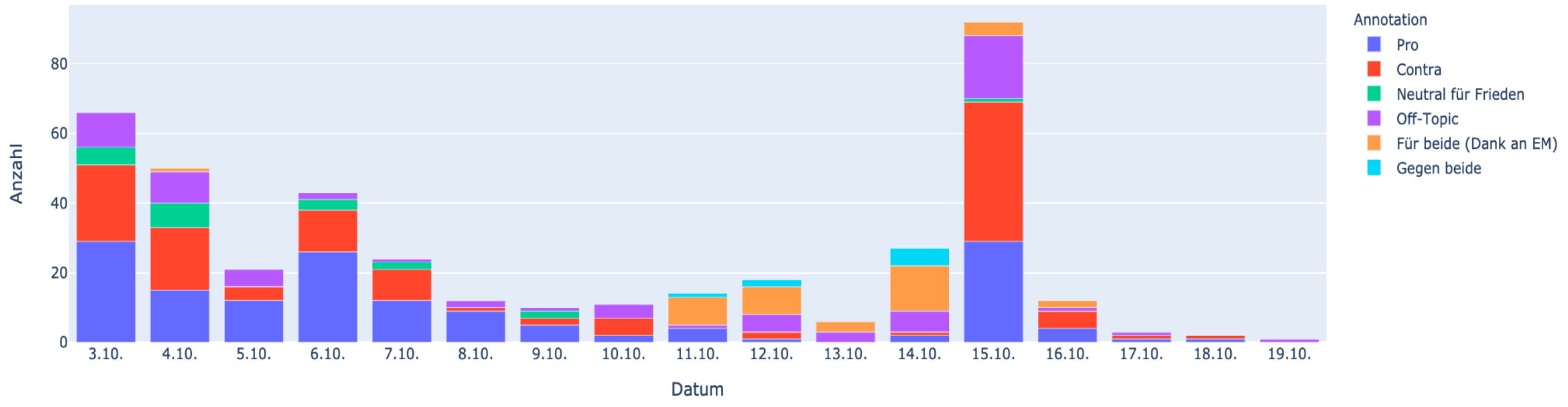

Abbildung 9: Verteilung der verschiedenen Gruppen über die Zeit im Elon-Musk-Shitstorm auf Twitter.

**Elon Musk – Twitter.** In Abb. 9 ist dargestellt, dass es zu Beginn vom 3. bis 9.10., eine neutrale Friedensseite gibt, die in der Mitte des Skandals verschwindet. Diese Gruppe positioniert sich weder für noch gegen Elon Musk. Am 11.10. erscheint dagegen die Gruppe »für beide, die Ukraine und Elon Musk«, die hauptsächlich von Ukrainer*innen repräsentiert wird, die Musk für seine Finanzierung des Starlink-Programms in der Ukraine danken. Gleichzeitig wächst die Zahl der Tweets gegen beide Seiten (Musk und die Ukraine), was auf russische Bots zurückzuführen sein könnte. Eines der Beispiele für Tweets, die diese Theorie stützen, ist der folgende, der zwar auf Englisch gepostet wurde, aber eindeutig eine russische Syntax aufweist und eine kremlfreundliche Erzählung mit antisemitischer Verschwörungstheorie und slawischer Supremazität enthält: »Here the only problem is that the person in charge of Ukraine is not Slavic, many are not, they are Jews with the face of capos who have revived Nazism, and now carry the eschatic, it is not Nazi, but Force Ukraine. Jews who are now Nazis«.

Es ist auch sichtbar, dass es zu Beginn mehr Befürworter*innen von Elon Musk gab (blau in Abb. 9), während zusammen mit den dankbaren Tweets nach dem 10. Oktober der Anteil der Personen, die gegen ihn waren, beträchtlich anstieg. Dies deutet auf das Hereinströmen von Ukrainer*innen von Telegram zu Twitter am 10.10. hin, ausgelöst durch die genannten Telegram-Posts.

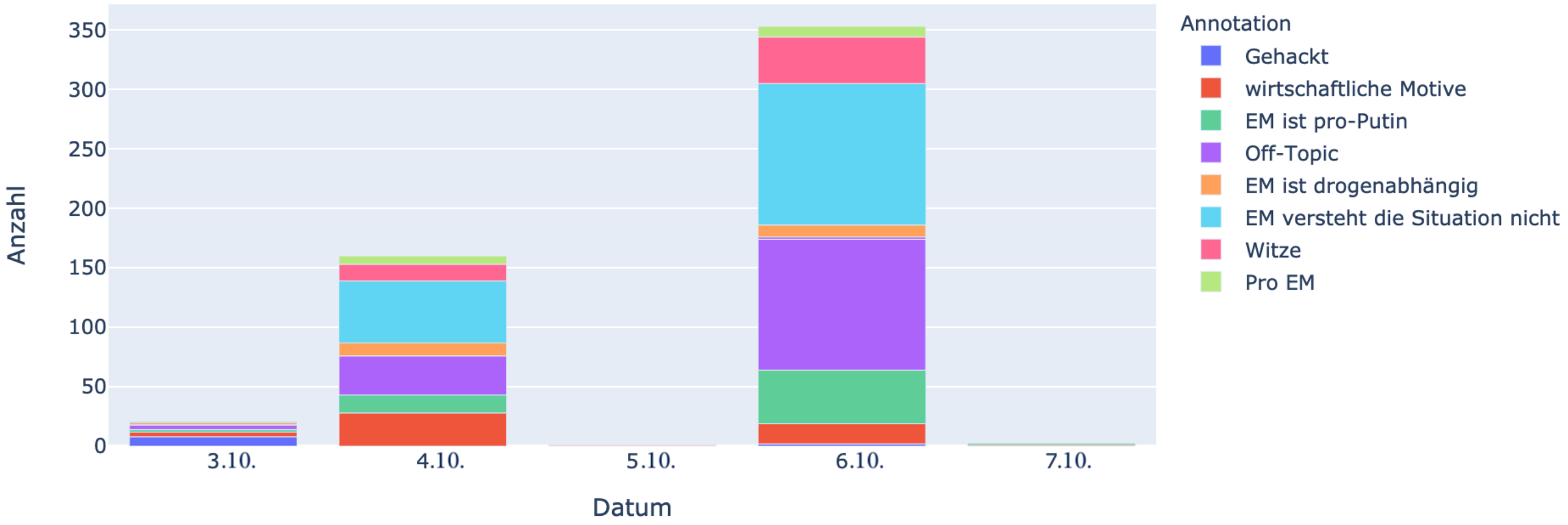

Abbildung 10: Verteilung der verschiedenen Gruppen über die Zeit im Elon-Musk-Shitstorm auf Telegram. Die ersten 7 sind alle gegen Elon Musk, haben aber unterschiedliche Erklärungen für sein Verhalten. Die letzte (grüne) Gruppe unterstützt ihn, weil er der Ukraine hilft.

Auf **Telegram** ist die Gruppe derer, die sich für Elon Musk entschuldigen, sehr klein, während die größte Gruppe behauptet, dass er die Situation und die Politik nicht versteht, dass er unmoralisch sei und sich daher nicht zu diesem Thema äußern sollte. Die zweitgrößte Gruppe ist diejenige, die behauptet, er verbreite Propaganda oder unterstütze Putin, und diese Gruppe wächst, solange Elon Musk weiter twittert. Die nächstgrößere Gruppe behauptet, dass es nur um Geld geht und er versuche, den enormen Rückgang der Tesla-Aktien zu verbergen, indem er die Aufmerksamkeit der Öffentlichkeit auf sich lenkt, aber ihr Anteil sinkt am 6.10. Ein großer Teil der Nachrichten enthält auch Witze, Anekdoten und allgemeine Heiterkeit, die sich immer gegen Elon Musk richtet.

Sprachlich unterscheiden sich die beiden Seiten der Interaktion. Im Elon-Musk-Skandal auf Twitter verwendet die Gruppe, die gegen ihn ist, mehr Konjunktionen, Adverbien, Begründungssätze, Zustandsverben, etwas mehr negative Emotionen und Wörter, die Wut und Traurigkeit ausdrücken. Im Gegensatz dazu verwenden die Befürworter mehr Emotionen der Erwartung (wie z. B. die Frage, wann die Ukraine endlich aufhören würde, stark von den USA finanziert zu werden), sowie etwas mehr Vertrauensemotionen und abstrakte Substantive. In Telegram macht die Gruppe der Befürworter von Elon Musk weniger als 1 % des Korpus aus. Es ist jedoch interessant, dass wir einen Unterschied zwischen russischsprachigen und ukrainischsprachigen Nutzer*innen in der Konversation auf Telegram beobachten konnten. Der russischsprachige Teil verwendet mehr Schlüsselwörter, die mit Geld, Vermögen und Aktien zu tun haben, während die ukrainischen Autor*innen mehr über die Ukraine, Russland und die Weltpolitik sprechen.

**eSports – Twitter.**

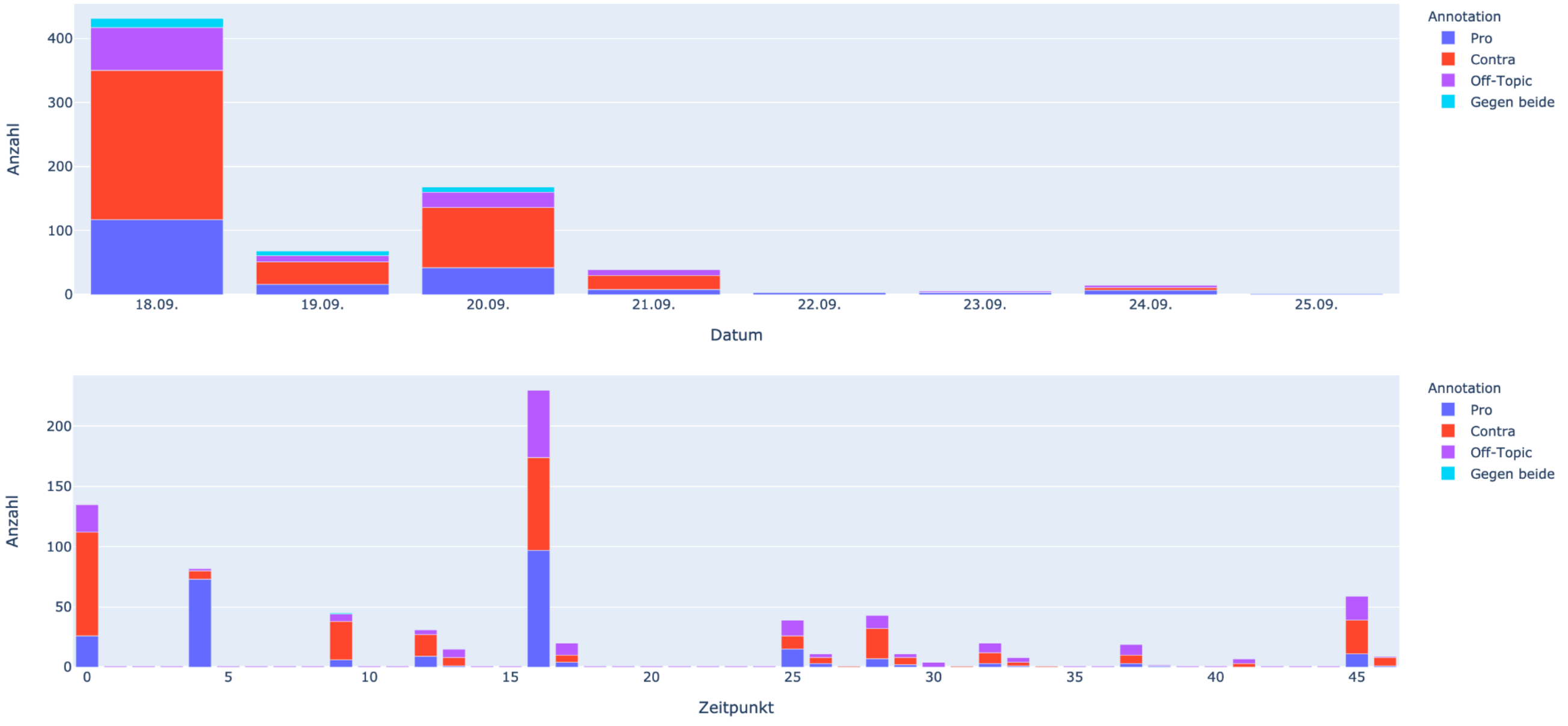

Abbildung 11: Verteilung der verschiedenen Gruppen über die Zeit im eSports-Shitstorm auf Twitter (oben) und Reddit (unten). Reddit wird in relativen Zeitpunkten angegeben.

Der Twitter-Skandal in der eSports-Gemeinde zeigt viel weniger Sympathie für die Zielperson. Anfangs sind nur 26 % auf seiner Seite und weniger als 10 % später, nach erfolglosem Schadensmanagement und möglicherweise mehr Beteiligten, die durch Reddit-Posts davon erfahren haben. Interessanterweise wächst die Gruppe, die gegen beide Seiten ist, prozentual leicht an. Dies hängt damit zusammen, dass sich der CEO öffentlich entschuldigt hat, was für diese Gruppe wie Verrat und Schwäche aussah und sie wollten, dass er sich »nicht dem Mob beugt«.

Wenn man die beiden Seiten in der eSports-Community auf Twitter betrachtet, verwendet die Opposition gegen den CEO von G2 viel mehr Personalpronomen und Konjunktionen und etwas mehr Adverbien. Sie ist auch die einzige Gruppe, die Abscheu ausdrückt, während das einzige Merkmal, das für die Befürworter etwas relevanter ist, abstrakte Substantive und der emotionale Wortschatz des Vertrauens sind.

Genaue Zeitstempel fehlen auf **Reddit**, aber wir beobachten ähnliche Spitzen wie auf Twitter am ersten Tag und am 20.9. Die Hauptgruppen sind hier die Unterstützer*innen von Ocelote und seine Gegner, da die Gruppe »gegen beide Seiten« für weniger als 1 % der Beiträge verantwortlich ist. Nach der anfänglichen, fast einhellig negativen Reaktion sind die Reddit-Nutzer*innen anfangs viel mitfühlender mit Ocelote, bis sich dies ändert und es mehr Gegner*innen gibt, obwohl er entlassen wurde. Eine Besonderheit dieses Subreddits ist auch, dass die Leute oft über viele andere Dinge diskutierten (in grün) und nicht nur über diese Situation, genau wie bei Telegram. Geschlossene, themenorientierte Gemeinschaften (wie bei Telegram oder Reddit) sind anscheinend viel weniger anfällig für Shitstorms, da diese Gemeinschaften offensichtlich viele andere Themen haben, während es bei Twitter in der Natur der Sache liegt, dass sich sehr unterschiedliche Teilnehmer*innen auf eine bestimmte, kontroverse Nachricht konzentrieren.

Auf Reddit scheinen die beiden sich gegenüberstehenden Gruppen jedoch eher ähnliche sprachliche Ausdrücke zu verwenden. Befürworter*innen verwenden wieder etwas mehr Adverbien und Konjunktionen, außerdem Zustandswörter, Nebensätze, Abscheu und interessanterweise die Konjunktion »aber«. Die Gruppe der Befürworter*innen verwendet hier mehr Personalpronomen.

## 4. Automatische sprachliche Differenzierung innerhalb des Shitstorms

Auf der Grundlage der oben vorgestellten chronologischen und gruppenspezifischen Analyse haben wir anhand der Daten in einem Pilotexperiment zwei Modelle trainiert, die in Zukunft bei der quantitativen Analyse von Shitstorms helfen könnten. Wir verwendeten das multilinguale BERT-Modell von HuggingFace, das auf mehr als 100 Sprachen trainiert wurde[10], und ein Klassifizierungsmodell auf der Grundlage von simpletransformers.

Zunächst haben wir die Nachrichten nach dem Zeitabschnitt klassifiziert, in dem sie im Shitstorm erscheinen. Wir unterscheiden drei zentrale Zeitabschnitte im Shitstorm: den intensiven Anfang, die Mitte bis zum letzten Peak, sowie das Ende. Wir erhielten eine Klassifikationsgenauigkeit für einzelne Beiträge von $F_1$=67 %. Dies zeigt, dass die Phasen des Shitstorms sich sprachlich unterscheiden. Mit mehr Feinabstimmung und größeren Trainingsdaten kann damit die Vorhersage der Phase, in der sich der Shitstorm gerade befindet, und somit auch die Vorhersage, wie weit er vom Ende entfernt ist, eine sehr erreichbare Realität sein.

Zweitens haben wir die Position des*der Beitragenden in drei Klassen eingeteilt: Unterstützer*innen, Gegner*innen und neutral gegenüber der Zielperson des Shitstorms. Wir erreichten ein geringeres Ergebnis bei dieser Aufgabe, mit $F_1$=62 %. Mit Blick auf die verschiedenen Gruppen der obigen Analyse, denken wir, dass diese niedrige Leistung größtenteils auf die Tatsache zurückzuführen ist, dass wir in den Klassen »für« und »gegen« eine Menge heterogener Unterklassen zusammengefasst haben. Unter anderem wurden beide Shitstorms gemeinsam betrachtet, so dass nur themenübergreifende Merkmale zum Zuge kommen können.

## 5. Diskussion

Online-Shitstorms entstehen in der Regel bei kontroversen Themen. Dies schafft die Voraussetzungen für heftige Auseinandersetzungen, bei denen die Nutzer*innen entweder gegensätzliche Positionen zu einem Thema verteidigen oder ablehnen. Ein wichtiger Faktor, der den Shitstorm wohl noch verstärkt, ist das Teilen von Empörung. Es hat sich gezeigt, dass Empörung und negative Emotionen Vorteile im Wettbewerb um Aufmerksamkeit haben, indem sie die Postenden dazu anregen, mehr Inhalte zu produzieren und zu konsumieren, oder indem sie Inhalte polarisieren. Dies erhöht sowohl die Dauer als auch die Intensität des Engagements, denn Inhalte, die Empörung hervorrufen, werden in der Regel häufiger angesehen und geteilt, so dass die Nutzer*innen immer wieder kommen. Bei empörenden Inhalten ist es daher wahrscheinlicher, dass sie sich viral verbreiten, indem sie eine emotionale Ansteckung über gegnerische ideologische Gemeinschaften hinweg erzeugen[11].

[10] Devlin, Jacob et al. (2022): »BERT: Pre-training of Deep Bidirectional Transformers for Language Understanding«, in: Proceedings of NAACL, Minneapolis, Minnesota, Internet: https://aclanthology.org/N19-1423. Zuletzt geprüft am: 30.11.2022. S. 4171–4186.

Die Netzwerkstruktur von Social-Media-Plattformen befriedigt unsere Vorliebe, zu einer »In-Group« zu gehören und sich in Cliquenstrukturen mit Gleichgesinnten zu mischen (dies wird als Homophilie bezeichnet). Diese Tendenz befriedigt unsere Suche nach Identität, die uns ein Gefühl der Zusammengehörigkeit vermittelt: Wir gleichen unsere Werte und Interessen mit denen anderer Menschen ab, die eine gemeinsame Identität und ähnliche Vorlieben haben. Doch unser Gefühl der Stammeszugehörigkeit macht uns auch anfällig für schädliche Narrative und Ideologien. Dies hat zur Folge, dass die Segregation in Gemeinschaften und die Konformität mit ihren Praktiken gefördert werden, was zu einer Polarisierung der Inhalte führen kann. Die Netzwerkstruktur spaltet die Gemeinschaften in In-Groups und Out-Groups. Dies kann zwei miteinander korrelierende Effekte haben: Entfremdung und Ausschluss der Stimme der wahrgenommenen Out-Groups und Schaffung eines Blasen-Effekts, bei dem die wahrgenommenen In-Groups sich gegenseitig in ihren Überzeugungen bestärken, da es keine Herausforderer und Andersdenkenden gibt. Kritisch anzumerken ist, dass diese Aufteilung der Inhalte nicht nur das Spektrum der zu hörenden Stimmen einschränkt, sondern auch zu einer Polarisierung der Meinungen führt, die zu immer extremeren Versionen dieser Meinungen führen.

In den betrachteten Daten ist die Cliquenbildung vor allem in den inhaltsbezogenen Plattformen Telegram und Reddit zu beobachten, wo eine Meinung stark überwiegt. Auf Twitter dagegen schaukelt sich die Empörung im Zusammenspiel der Unterstützer*innen und Gegner*innen hoch, die (teils nach expliziter Aufforderung) in einer öffentlichen Arena zusammenkommen. Auffallend ist, dass der Ablauf beider Shitstorms dem gleichen Schema folgt (Abb. 12).

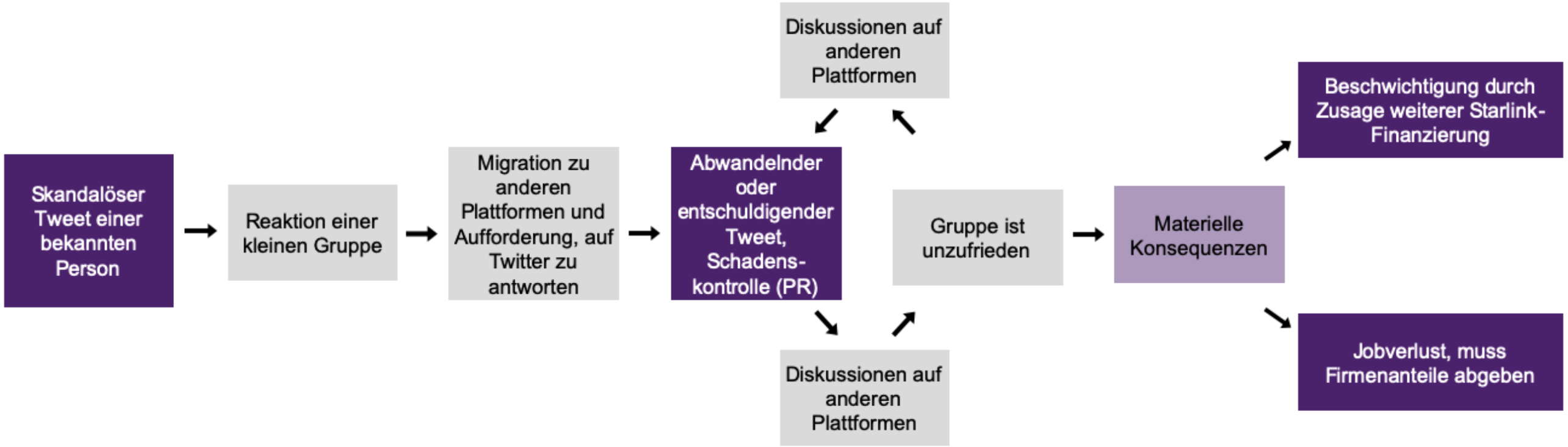

Abbildung 12: Schematischer zeitlicher Verlauf des plattformübergreifenden Shitstorms.

Aus beiden Beispielfällen leiten wir den schematischen zeitlichen Verlauf des plattformübergreifenden Shitstorms ab. Die Irritation wird zunächst durch einen skandalösen Beitrag auf Twitter ausgelöst, der von einer kleinen Gruppe Gegner*innen bemerkt wird, die diesen Inhalt in andere Plattformen tragen (wo die Gegner*innen thematisch organisiert sind). Durch eine beschwichtigende Reaktion der Zielperson auf die Kritik sind die Gegner*innen nicht zufriedengestellt und ziehen so mehr und mehr Beteiligte in die Diskussion: der sich verstärkende Shitstorm beginnt. Nach einigen Tagen kommt es zu materiellen Konsequenzen, die von Versprechungen (Fall Musk) bis zu Jobverlust und finanziellen Einbußen (Fall Ocelote) reichen können. Daraufhin schläft der Shitstorm schließlich ein bzw. wendet sich die Gemeinschaft anderen Themen zu.

[11] Popa-Wyatt, Mihaela (2023): Online Hate. Is Hate an Infectious Disease? Is Social Media a Promoter?; Popa-Wyatt, Mihaela (2022): Social media: A viral promoter of social ills?

## 6. Abbildungsverzeichnis

## 7. Literaturverzeichnis

# Anhang

## A. Elon-Musk-Shitstorm: Zeitliche Verteilung der Schlüsselwörter auf Twitter

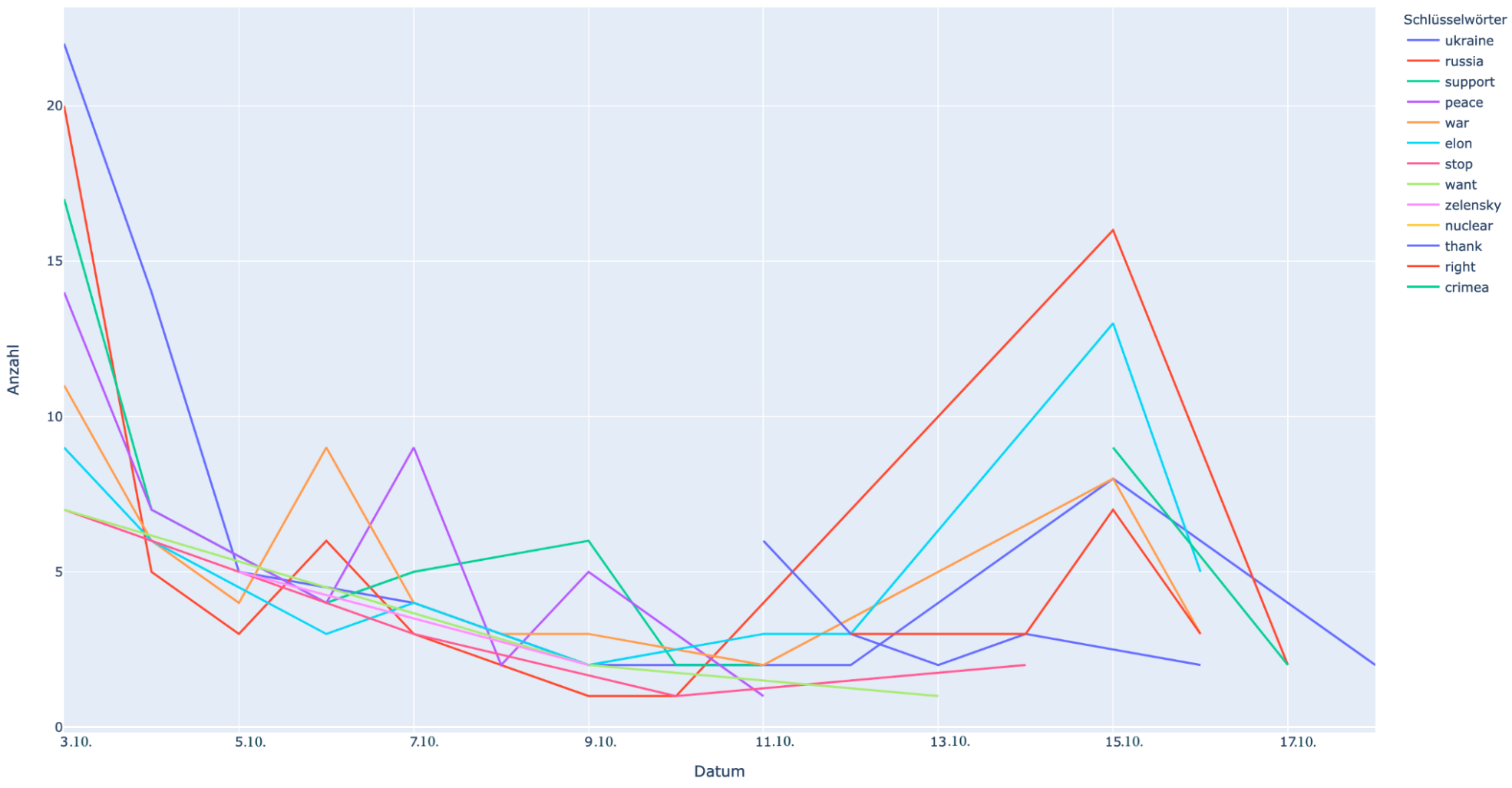

## B. eSports-Shitstorm: Zeitliche Verteilung der Schlüsselwörter auf Twitter

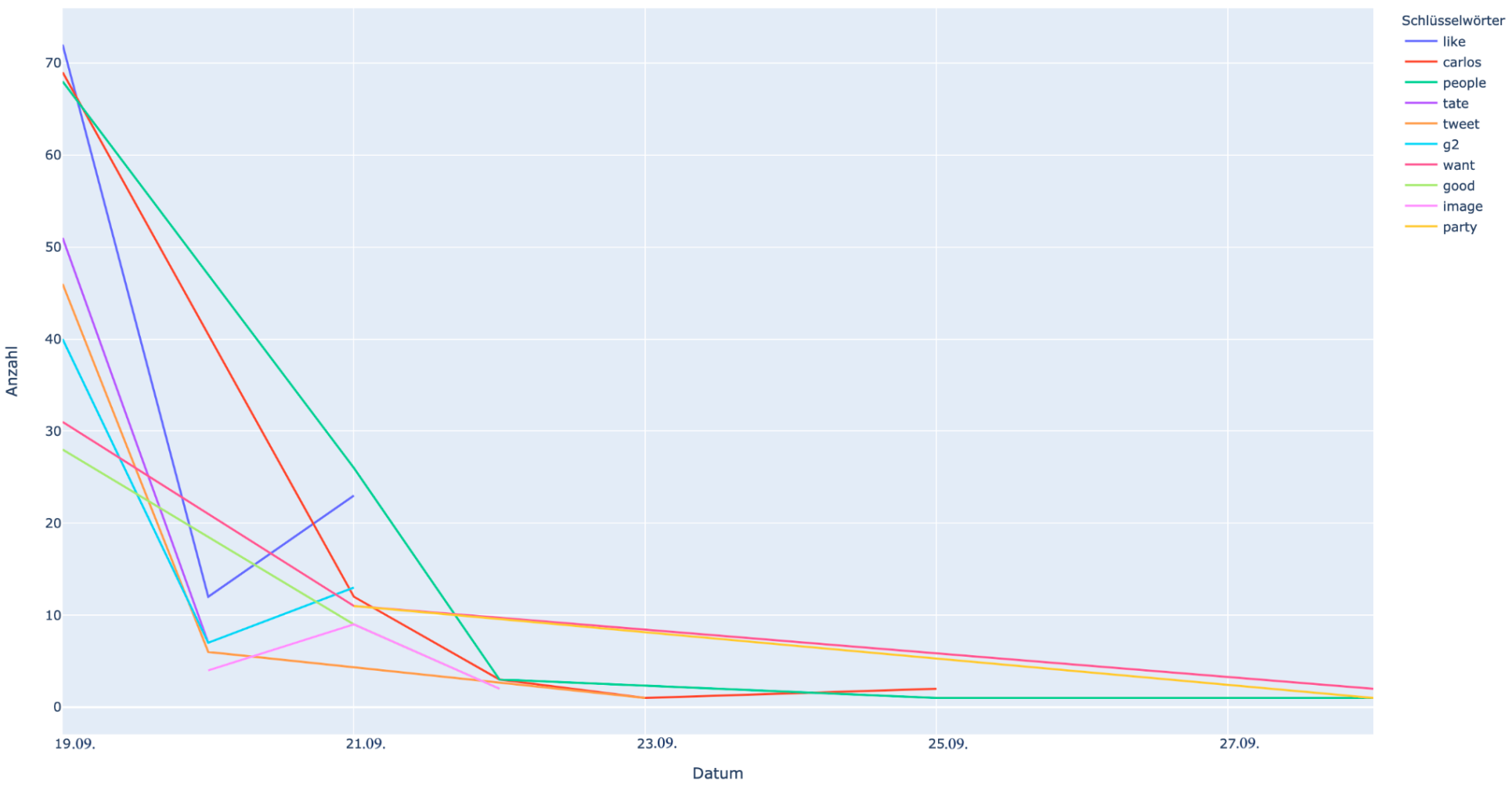

## C. eSports-Shitstorm: Zeitliche Verteilung der Schlüsselwörter auf Reddit

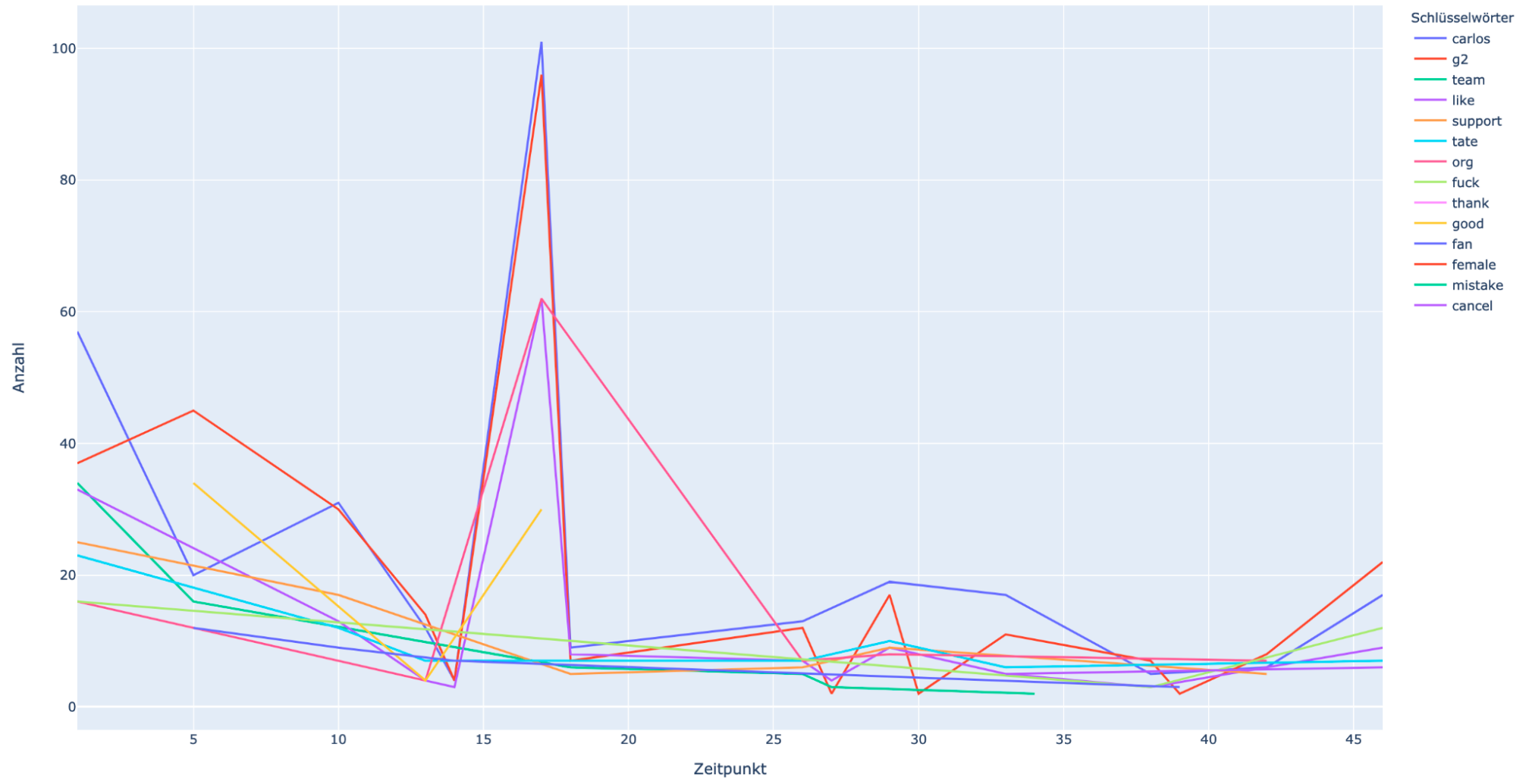

## D. Chronologische Analyse

Betrachtet man die Verteilung der Nachrichtenhäufigkeit über die Zeit, so kann man ein ähnliches Muster zwischen den zwei Shitstorms erkennen. Eine hohe Intensität zu Beginn, die kurz darauf abfällt und auf die zwei jeweils kleinere Spitzen folgen.

Wir haben mehrere linguistische Indikatoren betrachtet, um genauer zu untersuchen, wie sich die Sprache innerhalb des Shitstorms im Laufe der Zeit entwickelt. Wir haben jedes Korpus in 4 Zeitspannen unterteilt, wobei der Anfang der erste Peak ist, der zweite Teil der zweite Peak, der dritte Teil die Daten zwischen dem zweiten und dritten Peak und der letzte Peak das Ende. Aus jeder Nachricht extrahieren wir folgende linguistische Merkmale, die sich als sinnvoll zur Bestimmung opponierender Seiten in einer Debatte erwiesen haben[12]:

1.) Morphologische Merkmale: Anzahl der Adverbien, Adjektive, Verben, Eigennamen, Konjunktionen, Negationen, Anzahl der Komparative, Superlative, Personalpronomen, Passivformen;

2.) Syntaktische Merkmale: kontrastive Verwendung von »aber«, Anzahl der konzessiven, kausalen, und konsekutiven Teilsätze; Relativ-, Temporal- und Konditionalsätze.

3.) Interpunktionsmerkmale: Anführungszeichen, Fragezeichen;

4.) Oberflächliche Semantik: abstrakte Substantive, Modalverben, Zustandsverben, Behauptungen, Wörter mit hoher Modalität sowie Zählungen, die mit den Emotionen des NRC-Emotion-Lexikons übereinstimmen, wie z. B. Angst, Überraschung, Wut, Hoffnung, Ekel, Glück, Trauer, Vertrauen, negative und positive Gefühle, Wörter, die eine Behauptung oder eine Meinung ausdrücken.

5.) Durchschnittliche Satzlänge.

### D.2 Chronologie Elon-Musk-Shitstorm

**Twitter.** Eine erst abnehmende und dann wieder steigende Tendenz ist bei verschiedenen Merkmalen in der Twitter-Teilmenge Elon Musk für die Emotion der Überraschung, der Angst, sowie die

[12] Solopova V., Popescu O.I, Benzmüller C., & Landgraf T.: Automated multilingual detection of Pro-Kremlin propaganda in newspapers and Telegram posts. 2022.

Negationen und abstrakten Substantive zu beobachten. Die Emotion der Wut und die kausalen Nebensätze sind ebenfalls nur am Anfang und am Ende des Skandals vorhanden. Trauer und Überraschung sind nur zu Beginn des Shitstorms signifikant vorhanden, während Behauptungen hauptsächlich in der ersten und zweiten Periode gemacht werden. Schließlich wächst die Emotion der Hoffnung von Anfang an bis zur zweiten Periode, wo sie ansteigt und dann allmählich zum Ende hin abfällt.

**Telegram.** Das dritte Zeitfenster ist sehr signifikant für die Interaktion von Telegram (hier der 7. Oktober), mit einer etwas höheren Anzahl von Adjektiven, Konjunktionen, viel mehr Emotionen wie Hoffnung, Wut, Ekel, Überraschung, hohen Anteilen von Modalitätswörtern und Angst. Wut und Angst fehlen jedoch völlig zu Beginn des Sturms, dort ist auch der geringste Anteil an Verben des Konjunktivs zu finden. Kausale Nebensätze sind nur in der zweiten Periode und bei der höchsten Anzahl von Verben des Dativs vorhanden. Die fallend-steigende Tendenz ist bei den Eigennamen vorhanden, bei den Verben des Dativs ist sie jedoch umgekehrt. Die Anzahl der Personalpronomen steigt von Anfang an bis zur letzten Periode an, wo sie am höchsten ist. Die durchschnittliche Satzlänge ist viel höher als bei Twitter. Der erste Tweet von Elon Musk erhielt zum Zeitpunkt der Datenerhebung 59 Kommentare. In der höchsten Phase erreichte er ungefähr 90 Antworten. Gleichzeitig erfolgten in Telegram anfangs von einer viel weniger Reaktionen. Die Community in Telegram erreichte 150 und 350 Antworten in den beiden Spitzen (Peaks), entfernte sich dann aber offenbar schnell ganz vom Thema.

### D.2 Chronologie eSports-Shitstorm

**Reddit.** Im ersten und letzten Teil des Sturms werden vermehrt Adverbien, Adjektive, Personalpronomen, abstrakte Substantive, Verben im Konjunktiv und Konjunktionen verwendet. Zweckausdrücke und Personalpronomen der 1. Person Singular werden fast ausschließlich in den genannten Abschnitten verwendet, während Relativsätze fast nur im letzten Abschnitt vorkommen, ähnlich wie die Emotion des Ekels. Die Emotion des Vertrauens nimmt im letzten Zeitraum leicht zu, während die Emotion der Überraschung zu Beginn des Skandals fast nicht vorhanden ist, aber im weiteren Verlauf präsent ist.

**Twitter.** In der letzten Periode ist die durchschnittliche Satzlänge deutlich kürzer, während die Verwendung von Personalpronomen deutlich höher ist. Eine ähnliche Tendenz sehen wir bei abstrakten Substantiven und Zustandsverben. Behauptungen werden nur in der ersten Periode aufgestellt, wo auch der höchste Gebrauch von Konjunktionen zu beobachten ist (die dann nur in geringerem Maße in der dritten Periode erscheinen). Begründungssätze sind wiederum nur in der ersten und besonders häufig in der letzten Periode des Skandals zu finden, ebenso wie Glücksgefühle. Im Allgemeinen sind positive Emotionen im ersten Drittel und im letzten Zeitraum am stärksten vertreten, während sie im zweiten Zeitraum völlig fehlen. Wir können auch einen Anstieg der Verwendung von Eigennamen im dritten Zeitraum beobachten, in dem die Leute hauptsächlich über die verlorenen Sponsoren und die Heuchelei der Reaktion von Riot (dem großen Spielehersteller) diskutieren, der selbst eine Menge Anschuldigungen wegen sexueller Belästigung hat, aber eine so harte Strafe gegen den CEO von G2 verhängt hat. Wir können auch sehen, dass die Anzahl der verwendeten Adverbien von Tag 1 bis zum letzten Tag langsam abnimmt.